\documentclass[review,onefignum,onetabnum]{siamart190516}

\usepackage{lipsum}
\usepackage{amsfonts}
\usepackage{graphicx}
\usepackage{epstopdf}
\usepackage{algorithmic}
\ifpdf
  \DeclareGraphicsExtensions{.eps,.pdf,.png,.jpg}
\else
  \DeclareGraphicsExtensions{.eps}
\fi

\usepackage{enumitem}
\setlist[enumerate]{leftmargin=.5in}
\setlist[itemize]{leftmargin=.5in}
\usepackage{todonotes} 
\usepackage{amsmath,amssymb}
\usepackage{placeins}
\usepackage{cite}
\usepackage{float}

\newsiamremark{remark}{Remark}
\newsiamremark{hypothesis}{Hypothesis}
\crefname{hypothesis}{Hypothesis}{Hypotheses}
\newsiamthm{claim}{Claim}

\newcommand{\R}{\mathbb{R}}

\headers{A cortical-inspired model for Poggendorff-type illusions}{E. Baspinar, L. Calatroni, V. Franceschi and D. Prandi}

\title{A cortical-inspired sub-Riemannian model for Poggendorff-type visual illusions}

\author{Emre Baspinar\thanks{MathNeuro Team, INRIA Sophia Antipolis M{\'e}diterran{\'e}e, France  (\email{emre.baspinar@inria.fr}).}
\and Luca Calatroni\thanks{CNRS, UCA, INRIA, Morpheme, I3S, Sophia-Antipolis, France (\email{calatroni@i3s.unice.fr}).}
\and Valentina Franceschi\thanks{Dipartimento di Matematica Tullio Levi-Civita, Universit\`{a} di Padova  (\email{valentina.franceschi@unipd.it}).}
\and Dario Prandi\thanks{Universit\'{e} Paris-Saclay, CNRS, CentraleSup\'{e}lec, Laboratoire de Signaux et des Syst\`{e}mes, 91190, Gif-sur-Yvette, France  (\email{dario.prandi@centralesupelec.fr}).}}

\usepackage{amsopn}

\ifpdf
\hypersetup{
  pdftitle={A cortical-inspired sub-Riemannian model for Poggendorff-type visual illusions},
  pdfauthor={}
}
\fi

\usepackage[ruled, vlined]{algorithm2e}

\usepackage{ulem}

\usepackage{color}

\usepackage{mathtools}
\usepackage{environ}
\usepackage{amsmath}
\usepackage{amsfonts}
\usepackage{amssymb}
\usepackage[utf8]{inputenc}
\usepackage{caption}
\usepackage{subcaption}

\usepackage{todonotes}

\usepackage{placeins}
\usepackage[margin=3cm]{geometry}

\newcommand{\se}{\text{SE}(2)}
\newcommand{\sigmamu}{\sigma_\mu}

\graphicspath{{figures/}}

\begin{document}

\maketitle

\begin{abstract}
We consider Wilson-Cowan-type models for the mathematical description of orientation-dependent Poggendorff-like illusions. Our modelling improves two previously proposed cortical-inspired approaches embedding the sub-Riemannian heat kernel into the neuronal interaction term, in agreement with the intrinsically anisotropic functional architecture of V1 based on both local and lateral connections. For the numerical realization of both models, we consider standard gradient descent algorithms combined with Fourier-based approaches for the efficient computation of the sub-Laplacian evolution. Our  numerical results show that the use of the sub-Riemannian kernel allows to reproduce numerically visual misperceptions and inpainting-type biases in a stronger way in comparison with the previous approaches.
\end{abstract}


\section{Introduction}

\graphicspath{{figures/}}

The question of how we perceive the world around us has been an intriguing topic since the ancient times. 
For example, we can consider the philosophical debate around the concept of \emph{entelechy}, which started with the early studies of the Aristotelian school to answer this question, while, on the side  of phenomenology and on its relation to natural sciences, we can think of the theory started by Husserl. A well-known and accepted theory of perception is the one formulated within  Gestalt psychology  \cite{wertheimer1938laws, kohler1992gestal}.

The Gestalt psychology is a theory for understanding the principles underlying the configuration of local forms giving rise to a meaningful global perception. The main idea of the Gestalt psychology is that the mind constructs the whole by  grouping similar fragments rather than simply summing the fragments as if they were all different. In terms of visual perception, such similar fragments correspond to point stimuli with same (or very close) valued features of the same type. As an enlightening example from vision science, we tend to group the same colored objects in an image and to perceive them as an ensemble rather than the objects with different colors. There have been many psychophysical studies which attempted to provide quantitative parameters describing the tendencies of the mind in visual perception based on Gestalt psychology. A particularly important one is the pioneering work of Field \textit{et. al.} \cite{field1993contour}, where the authors proposed a representation, called \emph{association field}, modelling specific Gestalt principles. Furthermore, they also showed that it is more likely that the brain perceives together fragments which are similarly oriented and aligned along a curvilinear path than the ones rapidly changing orientations.

\begin{figure}[th]
\begin{subfigure}[b]{0.4\textwidth}
\centering
 \includegraphics[width = \linewidth]{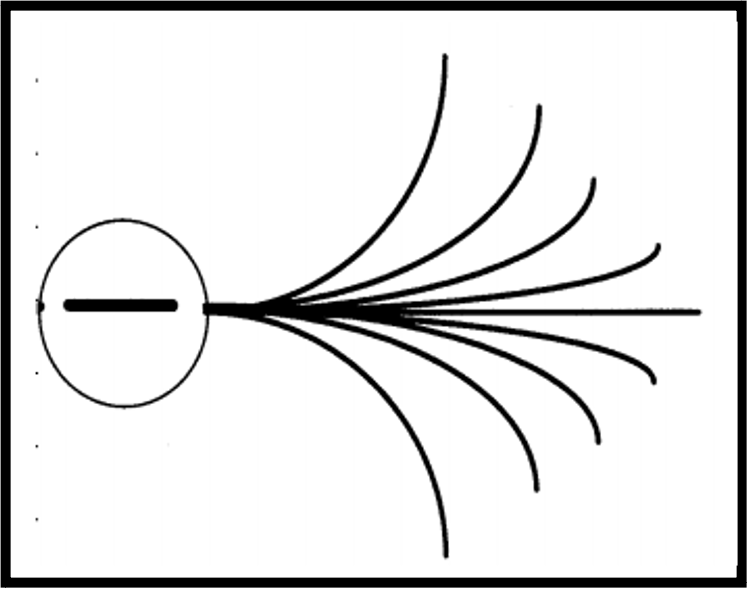}
\caption{Association fields}
\label{fig:association1}
\end{subfigure}
\hspace{2.5cm}
\begin{subfigure}[b]{0.4\textwidth}
\centering
\includegraphics[ width =\linewidth]{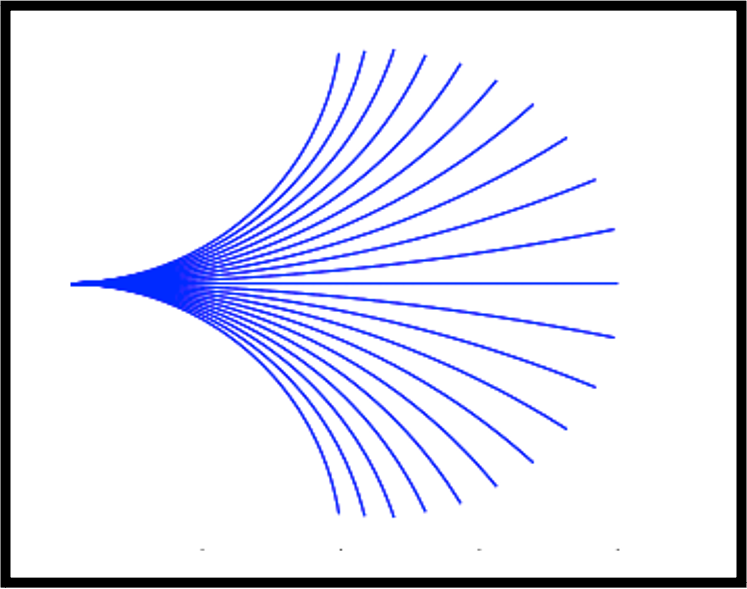}
\caption{Projected horizontal integral curves}
\label{fig:association2}
\end{subfigure}
\caption{Projections of horizontal integral curves approximate the association fields from the experiment of Field, Hayes and Hess \cite{field1993contour}. They are generated by the sub-Riemannian model geometry proposed by Citti and Sarti \cite{citti2006cortical}. Figures are adapted from \cite{field1993contour} and \cite{citti2006cortical}.}
\label{fig:association}
\end{figure}

The presented model for neuronal activity is a geometrical abstraction of the orientation-sensitive V1 hypercolumnar architecture observed by Hubel and Wiesel \cite{hubel1959receptive, hubel1962receptive, hubel1963shape}. This  abstraction generates a good \emph{phenomenological} approximation of the V1 neuronal connections existing in the hypercolumnar architecture as reported by Bosking  \textit{et.\ al.} \cite{bosking1997orientation}. In this framework, the corresponding projections of the neuronal connections  in V1 onto a 2D image plane are considered to be the association fields described above and the neuronal connections are modeled as the horizontal integral curves generated by the model geometry. The projections of such horizontal integral curves were shown to produce a close approximation of the association fields, see Figure~\ref{fig:association}. For this reason, the approach considered by Citti, Petitot and Sarti and used in this work is referred to be cortically-inspired.

We remark that the presented model for neural activity is a phenomenological model which provides a mathematical understanding of early perceptual mechanisms at the cortical level by starting from very structure of receptive profiles. Nevertheless, it has been very useful for many image-processing applications, see for example, \cite{BCGPR18, zhangNumerical2016}.

In this work, we follow this approach for a better understanding of the visual perception biases due to visual distortions often referred to as visual illusions.
Visual illusions are described as the mismatches between the reality and its visual perception. They result either from a neural conditioning introduced by external agents such as drugs, microorganisms, tumors \cite{levi1990visual,gaillard2003persisting} or from self-inducing mechanisms evoking visual distortions via proper neural functionality applied on specific stimulus \cite{hine1995illusion,prinzmetal2001ponzo}. The latter type of illusions are due to the effects of neurological and biological constraints on the visual system, \cite{purves2008visual}.

\begin{figure}[htb!]
\centering
\includegraphics[height=5cm]{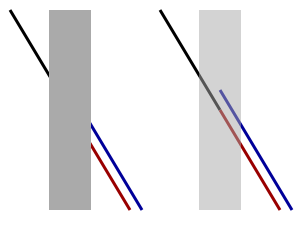}
\caption{The original Poggendorff illusion: The red colored line is aligned  with the black line although the blue one is falsely perceived as its continuation. Source: Wikipedia.}
\label{fig:Poggendorff}
\end{figure}

In this work, we focus on illusions induced by contrast induction and orientation misalignments, with a particular focus on the well-known Poggendorff illusion and its variations, see Figure \ref{fig:Poggendorff}. This is a geometrical optical illusion \cite{westheimer2008illusions,Weintraub1971} in which a misaligned oblique perception is induced by the presence of a central bar \cite{day1976components}.

\subsection{The functional architecture of the primary visual cortex } \label{sec:V1}

It has been known since the celebrated experiments of Hubel and Wiesel \cite{hubel1959receptive, hubel1962receptive, hubel1963shape} that neurons (simple cells) in the primary visual cortex (V1) perform boundary (hence orientation) detection and propagate their activations through the cortical connectivity, in accordance with the psychophysical results of Fields and Hayes \cite{field1993contour}. Hubel and Wiesel showed that simple cells have a spatial arrangement based on the so-called \emph{hypercolumns} in V1. In this arrangement, simple cells which are sensitive to different orientations at the same retinal location are found in the same vertical column constructed on the cortical surface. Adjacent columns contain the simple cells which are sensitive to close positions.

Several models have been proposed to describe the functional architecture of V1 and the neural connectivity within it. Koenderink et.\ al.\ \cite{koenderink1984structure, koenderink1987representation} focused on differential geometric approaches to study the visual space where they modelled the invariance of simple cells with respect to suitable symmetries in terms of a family of Gaussian functions. Hoffman \cite{hoffman1970higher, hoffman1989visual} provided the basic framework of vision models by interpreting the hypercolumn architecture of V1 as a fiber bundle. Following a similar reasoning, 
Petitot and Tondut \cite{petitot1999vers} further developed this modelling, providing a new model, coherent both with the structure of orientation sensitive simple cells and the long range neural connectivity between V1 simple cells. In their model, they first observed that the simple cell orientation selectivity induces a contact geometry (associated to the first Heisenberg group) rendered by the fibers of orientations. Moreover, they showed that a specific family of curves found via a constrained minimization approach in the contact geometry fits the aforementioned association fields reported by Field et.\ al.\ \cite{field1993contour}. 
In \cite{citti2006cortical, citti2014neuromathematics}, Citti and Sarti further developed the model of Petitot and Tondut, by introducing a group based approach, then refined by Boscain, Gauthier \textit{et al.} \cite{BDGR12, boscainHypoelliptic2014}. See also the monograph \cite{PGbook}.
The so-called Citti-Petitot-Sarti (CPS) model exploits the natural sub-Riemannian (sR) structure of the group of rotations and translations $\se$ as the V1 model geometry. 

In this framework, simple cells are modelled as points of the three-dimensional group $\mathcal M = \R^2\times \mathbb P^1$. Here, $\mathbb P^1$ is the projective line, obtained by identifying antipodal points in $\mathbb S^1$. The response of simple cells to the two-dimensional visual stimuli is identified by lifting them to $\mathcal M$ via a Gabor wavelet transform. 
Neural connectivity is then modelled in terms of \emph{horizontal integral curves} given by the natural sub-Riemannian structure of $\mathcal M$.
Activity propagation along neural connections can further be modelled in terms of diffusion and transport processes along the horizontal integral curves. 

In recent years, the CPS model have been exploited as a framework for several cortical-inspired image processing problems by various researchers. We mention the large corpus of literature by Duits \textit{et al.}, see e.g., \cite{duits2009line, duits2010left1, duits2010left2} and the state-of-the-art image inpainting and image recognition algorithms developed by Boscain, Gauthier, \textit{et al.} \cite{bohiFourier2016,BCGPR18}. 
%
Some extensions of the CPS model geometry and its applications to other  image processing problems can be found in \cite{bekkers2014multi, barbieri2014cortical, citti2016sub, baspinar2018geometric, janssen2018design, franceschiello2018neuromathematical,lafarge2020roto, baspinar2020sub}.

\subsection{Mean-field neural dynamics \& visual illusions}

Understanding neural behaviors is in general a very challenging task. Reliable responses to stimuli are typically measured at the level of population assemblies comprised by a large number of coupled cells. This motivates to reduce, whenever possible, the dynamics of a neuronal population to a neuronal mean-field model which describes large-scale dynamics of the population as the number of neurons goes to infinity. These mean-field models, inspired by the pioneering work of Wilson and Cowan \cite{wilson1972excitatory, wilson1973mathematical}, and Amari \cite{amari1977dynamics}, are low dimensional in comparison to their corresponding ones based on large-scale population networks. Yet, they capture the same dynamics underlying the population behaviors. 

In the framework of the CPS model for V1 discussed above, several mathematical models were proposed to describe the neural activity propagation favoring the creation of visual illusions, including Poggendorff type illusions. 
In \cite{franceschiello2018neuromathematical}, for instance, illusions are identified with suitable strain tensors, responsible of the perceived displacement from the grey levels of the original image. In \cite{franceschiello2019geometrical}, illusory patterns are identified by a suitable modulation of the geometry of $\se=\R^2\times S^1$ and computed as the associated geodesics via the fast-marching algorithm.

In \cite{bertalmio2020cortical,bertalmio2020visual,SSVMproceeding2019}, a variant of the Wilson-Cowan (WC) model based on a variational principle and adapted to the $\mathcal M$ geometry of V1 was employed to model the neuronal activity and generate illusory patterns for different illusion types. The modelling considered in these works is strongly inspired by the integro-differential model firstly studied in \cite{bertalmio2007perceptual} for perception-inspired Local Histogram Equalization (LHE) techniques and later applied in a series of work, see, e.g., \cite{bertalmio2009implementing,BertalmioFrontiers2014} for the study of contrast and assimilation phenomena. By further incorporating a cortical-inspired modelling, the authors showed in \cite{bertalmio2020cortical,bertalmio2020visual,SSVMproceeding2019} that cortical LHE models are able to replicate visual misperceptions induced not only by local contrast changes, but also by orientation-induced biases similar as the ones in Figure \ref{fig:Poggendorff}. Interestingly, the cortical LHE model \cite{bertalmio2020cortical,bertalmio2020visual,SSVMproceeding2019}  was further shown to outperform both standard and cortical-inspired WC models and rigorously shown to correspond to the minimization of a variational energy, which suggests better efficient representation properties \textcolor{blue}{\cite{Attneave1954,Barlow1961}}. One major limitation in the modelling considered in these works is the use of neuronal interaction kernels (essentially, isotropic 3D Gaussian) which are not compatible with the natural sub-Riemannian structure of V1 proposed in the CPS model.

\subsection{Main contributions}

In this work, we encode the sub-Riemannian structure of V1 into both WC and LHE models by using a sub-Laplacian procedure associated with the geometry of the space $\mathcal M$ described in Section \ref{sec:V1}.  Similarly as in \cite{bertalmio2020cortical,bertalmio2020visual,SSVMproceeding2019}, the lifting procedure associating to a given two dimensional image the corresponding neuronal response in $\mathcal M$  is performed by means of all-scale cake wavelets, introduced in \cite{duits2005perceptual,duits2007invertible}. A suitable gradient-descent algorithm is applied to compute the stationary states of the neural models. 

Within this framework, we study the family of Poggendorf visual illusions induced by local contrast and orientation alignment of the objects in the input image. In particular, we aim to reproduce such illusions by the proposed models in a way which is qualitatively consistent with the psychophysical experience. 

Our findings show that it is possible to reproduce Poggendorff-type illusions by both the sR cortical-inspired WC and LHE models. This, compared with the results in \cite{bertalmio2020cortical, bertalmio2020visual} where the cortical WC model endowed with a Riemannian (isotropic) 3D kernel was shown to fail to reproduce Poggendoff-type illusions, shows that adding the natural sub-Laplacian procedure in the computation of the flows improves the capability of those cortical-inspired models in terms of reproducing orientation-dependent visual illusions.

\section{Cortical-inspired modelling}   \label{sec:cortical_modelling}

In this section we recall the fundamental features of CPS models. The theoretical criterion underpinning the model relies on the so-called neurogeometrical approach introduced in \cite{petitot1999vers, citti2006cortical,sarti2008symplectic}. According to this model, the functional architecture of V1 is based on the geometrical structure inspired by the neural connectivity in V1.

\subsection{Receptive profiles}
A simple cell is characterized by its \emph{receptive field}, which is defined
as the domain of the retina to which the simple cell is sensitive. Once a receptive field is stimulated, the corresponding retinal cells generates spikes which are transmitted to V1 simple cells via retino-geniculo-cortical paths. 

The response function of each simple cell to a spike is called \emph{receptive profile} (RP), and denoted by $\psi_{(\zeta,\theta)}:Q\to \mathbb C$. It is basically the impulse response function of a V1 simple cell. Conceptually it is the measurement of the response of the corresponding V1 simple cell to a stimulus at a point\footnote{Note that we omit the coordinate maps between the image plane and retina surface, and the retinocortical map from the retina surface to the cortical surface. In other words, we assume that the image plane and the retinal surface are identical and denote both by $Q\subset\R^2$.} $\zeta=(x,y)\in Q$. 

In this study we assume the response of simple cells to be linear. That is, for a given visual stimulus $f:Q\to \R$ we assume the response of the simple cell at V1 coordinates $(\zeta,\theta)$ to be
\begin{equation}\label{eq:firstEq}
    a_0(\zeta,\theta) = \langle f, \psi_{(\zeta,\theta)} \rangle_{L^2(Q)} 
    =\int_Q \psi_{(\zeta,\theta)}(u)f(u)\,du.
\end{equation}
This procedure defines the cortical stimulus $a_0:\mathcal M\to \mathbb C$ associated with the image $f$.
We note that receptive field models consisting of cascades of linear filters and static non-linearities, although not perfect, may be more adequate to account for responses to stimuli \cite{koenderink1987representation, bekkers2018roto, lindeberg2013computational}. Several mechanisms such as, e.g., response normalization, gain controls, cross-orientation suppression, or intra-cortical modulation, might intervene to change radically the shape of the profile. Therefore, the above static and linear model for the receptive profiles should be considered as a first approximation of the complex behavior of a real dynamic receptive profile, which cannot be perfectly described by static wavelet frames. 

Regarding the form of the RP, in  \cite{citti2006cortical}, a simplified basis of Gabor functions were proposed as good candidates for modelling the position-orientation sensitive receptive profiles for neuro-physiological reasons \cite{daugman1985uncertainty, barbieri2012uncertainty}. This basis has then been extended to take into account additional features such as scale \cite{sarti2008symplectic}, velocity \cite{barbieri2014cortical}, and frequency-phase \cite{baspinar2020sub}. On the other hand, Duits et al. \cite{duits2007invertible} proposed so-called \emph{cake kernels} as a good alternative to Gabor functions, and showed that cake kernels were adequate for obtaining simple cell output responses which were used to perform certain image processing tasks such as image enhancement and completion based on sR diffusion processes . 

In this study, we employ cake kernels as the models of position-orientation RPs obtaining the initial simple cell output responses to an input image, and we  use the V1 model geometry $\mathcal M$ to represent the output responses. We model the activity propagation along the neural connectivity by using the combination of a diffusion process based on the natural sub-Laplacian and a Wilson-Cowan type integro-differential system. 

\subsection{Horizontal connectivity and sub-Riemannian diffusion}
\label{sec:connectivity}

Neurons in V1 present two type of connections: local and lateral. Local connections connect neurons belonging to the same hypercolumn. On the other hand, lateral connections account for the connectivity between neurons belonging to different hypercolums, but along a specific direction. In the CPS model these are represented\footnote{This expression does not yield smooth vector fields on $\mathcal M$. Indeed, e.g., $X_1(\zeta, 0)=-X_1(\zeta,\pi)$ despite that $0$ and $\pi$ are identified in $\mathbb P^1$. Although in the present application such difference is inconsequential, since we are only interested in the direction (which is smooth) and not in the orientation, this problem can be solved by defining $X_1$ in an appropriate atlas for $\mathcal M$ \cite{BDGR12}.} by the vector fields
\begin{equation}\label{eq:horizontal_VFs}
    X_1 = \cos\theta\partial_x + \sin\theta\partial_y, \qquad X_2=\partial_\theta.
\end{equation}

The above observation yield to the modelling of the dynamic of the neuronal excitation $\{Z_t\}_{t\ge_0}$ starting from a neuron $(\zeta,\theta)$ via the following stochastic differential equation
\begin{equation}
    dZ_t = X_1 du_t + X_2 dv_t, \qquad Z_0 = (\zeta,\theta),
\end{equation}
where $u_t$ and $v_t$ are two one-dimensional independent Wiener processes.
As a consequence, in \cite{BDGR12} the cortical stimulus $a_0$ induced by a visual stimulus $f_0$ is assumed to evolve according to the Fokker-Planck equation
\begin{equation}\label{eq:heat-sr}
    \partial_t \psi = \mathcal L \psi, \qquad \mathcal L = X_1^2+\beta^2X_2^2.
\end{equation}
Here, $\beta>0$ is a constant encoding the unit coherency between the spatial and orientation dimensions.

The operator $\mathcal L$ is the sub-Laplacian associated to the sub-Riemannian structure on $\mathcal M$ with orthonormal frame $\{X_1,X_2\}$, as presented in \cite{citti2006cortical, BDGR12}. It is worth mentioning that this operator is not elliptic, since $\{X_1,X_2\}$ is not a basis of $T\mathcal M$. However, $\operatorname{span}\{X_1,X_2,[X_1,X_2]\}=T\mathcal M$. Hence, $\{X_1,X_2\}$ satisfies the Hörmander condition and $\mathcal L$ is an hypoelliptic operator \cite{hormander1967hypoelliptic} which models the activity propagation between neurons in V1 as the diffusion concentrated to a neighborhood along the (horizontal) integral curves of $X_1$ and $X_2$. 

A direct consequence of hypoellipticity is the existence of a smooth kernel for \eqref{eq:heat-sr}. That is, there exists a function $(t,\xi,\nu)\in \R_+\times\mathcal M\times\mathcal M\mapsto k_t(\xi,\nu)$ such that the solution of \eqref{eq:heat-sr} with initial datum $a_0$ reads
\begin{equation}\label{eq:evolution}
    \psi(t, \xi) = e^{t\mathcal L} a_0(\xi) = \int_{\mathcal M} k_t(\xi,\nu)a_0(\nu)\,d\nu.
\end{equation}
An analytic expression for $k_t$ can be derived in terms of Mathieu functions \cite{duits2010left1, zhangNumerical2016}. This expression is however cumbersome to manipulate, and it is usually more efficient to resort to different schemes for the numerical implementation of \eqref{eq:heat-sr}, see, e.g.,~Section~\ref{sec:Algorithms}.

\subsection{Reconstruction on the retinal plane}

Activity propagation evolves the lifted visual stimulus in time.  In order to obtain a meaningful result,  which is represented on a 2-dim image plane, we have to transform back the evolved lifted image to the 2-dim image plane. We achieve this by using the projection given by
\begin{equation}\label{eq:reconstruction}
f(\zeta,T) = \int_0^\pi a(\zeta,\theta,T)\,d\theta,
\end{equation}
where $f: \R^2\times (0,T]\rightarrow \R$ and $0<T<\infty$ denote the processed image and the final time of the evolution, respectively. One easily checks that this formula yields $f(\cdot, 0) = f_0$ under the assumption
\begin{equation}
    \int_{0}^\pi \psi_{\xi,\theta}(u)\,d\theta = 1.
\end{equation}

\section{Describing neuronal activity via Wilson-Cowan-type models}  \label{sec:wc_models}

In neurophysiological experiments, reliable neural responses to visual stimuli are generally observed at the neuronal population level: the information processing and the response produced are obtained by integrating the individual dynamics of the neurons interacting within the population. Modelling neuronal populations can be done via coupled differential systems (networks) consisting of a large number of equations, and the average behavior can in principle be used to represent the population behavior. This requires high computational power and the use of challenging analytical approaches due to the high dimension of the network. A different mesoscopic approach consists in  considering the average network behavior as the  number of neurons in the network is let to infinity. The asymptotic limit of the network can thus be written in terms of the probability distribution (density) of the state variables. This asymptotic limit is the so-called mean-field limit. It has been successfully used as a reference framework in several papers, see, e.g., \cite{Faugeras2009,Bressloff2002,WCreview2009} and will also be the approach considered in this work.

\subsection{Wilson-Cowan (WC) model}
Let $a(\zeta,\theta, t)$ denote the evolving activity of the neuronal population located at $\zeta\in\mathbb{R}^2$ and sensitive to the orientation $\theta\in\mathbb{P}^1$ at time $t\in(0,T]$. By using the shorthand notation $\xi=(\zeta,\theta),\; \eta=(\nu, \phi)\in \mathcal M$, the Wilson-Cowan (WC) model on $Q\subset\mathbb{R}^2$ can be written as follows:
\begin{equation}\label{eq:WCRef}
\partial_t a(\xi,t) = -(1+\lambda) a(\xi,t)+ \frac{1}{2M}\int_{Q\times [0,\pi)}\omega_{\xi}(\eta)\sigma\Big( a(\eta,t) \Big )d\eta + \lambda a_0(\xi)+\mu(\xi).
\end{equation}
Here $\mu:Q\to \mathbb{R}$ is a smoothed version of the simple cell output response $a_0$ via a Gaussian filtering, while parameters $\lambda>$ and $M>0$ are fixed positive constants. Following the standard formulation of WC models studied, e.g., in \cite{Faugeras2009,sarti2015constitution} we have that the role of the time-independent external stimulus $h:Q\times[0,\pi)\to \mathbb{R}$ is played here by $h(\xi):= \lambda a_0(\xi)+\mu(\xi)$   while model parameters can be set as $\beta:=1+\lambda$ and $\nu:=1/2M$. The function $\sigma:\R\to [-1, 1]$ stands for a nonlinear saturation function, which we choose as the sigmoid:
\begin{equation}\label{eq:sigma}
\sigma(r) := -\min\Big(1, \max(\alpha (r-1/2),\,-1)   \Big),\quad \alpha>1.
\end{equation}
The connectivity kernel $\omega_{\xi}$ models the interaction between neurons in $\mathcal M$. Its definition should thus take into account the different type of interactions happening between connected neurons in V1, e.g. it should model at the same time both local and lateral connections via the sub-Riemannian diffusion described in Section \ref{sec:connectivity}.

In \cite{bertalmio2020cortical,bertalmio2020visual} the authors showed that \eqref{eq:WCRef} does not arise from a variational principle. That is, it there exists no energy function $E:L^2(\mathcal M)\to \mathbb R$ such that \eqref{eq:WCRef} can be recast as the problem
\begin{equation}\label{eq:variationalFormula}
\partial_t a (\xi,t)= -\nabla E (a(\xi,t)),\qquad a(\xi,0)=a_0 = L f_0.
\end{equation}
Under this formulation, stationary states $a^*$ of \eqref{eq:WCRef} are (local) minima of $E$.

The interest of considering an evolution model following a variational principle in the sense \eqref{eq:variationalFormula} is given by its connection with the optimization-based approaches considered in \cite{Olshausen2000} to describe the efficient coding problem as an energy minimization problem which involves natural image statistics and biological constraints which force the final solution to show the least possible redundancy.
Under this interpretation, the non-variational model \eqref{eq:WCRef} is suboptimal in reducing redundant information in visual stimuli, see \cite[Section 2.1]{bertalmio2020cortical} for more details. 

\subsection{Local Histogram Equalization (LHE) model}

In order to build a model which complies with the efficient neural coding described above, in \cite{bertalmio2020cortical,bertalmio2020visual}, the authors showed that \eqref{eq:WCRef} can be transformed into a variational problem by replacing the term $\sigma(a(\eta,t))$ with $\hat{\sigma}(a(\xi, t)- a(\eta,t))$ for a suitable choice of the nonlinear sigmoid function $\hat{\sigma}$, thus enforcing non-linear activations on local contrast rather than on local activity. The corresponding model reads:
\begin{equation}\label{eq:LHERef}
\partial_t a(\xi,t) = -(1+\lambda)a(\xi,t) + \frac{1}{2M}\int_{Q\times [0,\pi)}\omega_{\xi}(\eta)\hat{\sigma}\Big( a(\xi,t) - a(\eta, t) \Big )d\eta + \lambda a_0(\xi)+\mu(\xi).
\end{equation}
 where $\hat{\sigma}(r) := -\sigma ( r+1/2 )$, and $\sigma$ as in \eqref{eq:sigma}. 
This model has been first introduced in \cite{bertalmio2007perceptual} as a variational reformulation of the Local Histogram Equalization (LHE) procedure for RGB images. The corresponding energy $E:L^2(\mathcal M)\to \mathbb R$ for which  \eqref{eq:variationalFormula} holds is:
\begin{multline}\label{eq:lhe-energy}
    E(a) =  
    \frac{\lambda}{2} \int_{Q\times[0,\pi)}|a(\xi)- a_0(\xi) |^2\,d\xi  
    + \frac{1}{2} \int_{Q\times[0,\pi)}|a(\xi)- \mu(\xi) |^2\,d\xi\\
    +\frac1{2M}\int_{Q\times[0,\pi)}\int_{Q\times[0,\pi)}\omega_\xi(\eta) \Sigma\left( a(\xi)-a(\eta)\right)\,d\xi\,d\eta
    ,
\end{multline}
where $\Sigma:\mathbb R\to \mathbb R$ is any (even) primitive function for $\hat{\sigma}$.

As it is clear from \eqref{eq:lhe-energy},  the local histogram equalization properties of the model are due here to the activation averaging which is localized by the kernel  $\omega_\xi$, which should thus be adapted to the natural geometry of $\mathcal M$ (see Section \ref{sec:kernel} for a more detailed discussion).

\subsection{A sub-Riemannian choice of the interaction kernel $\omega_\xi$}  \label{sec:kernel}

In \eqref{eq:WCRef} and \eqref{eq:LHERef}, the geometric structure of the underlying space $\mathcal{M}$ is captured by the connectivity kernel $\omega_\xi$, which characterizes the activity propagation along neural connections in V1. In \cite{bertalmio2020cortical, bertalmio2020visual}, simple 3-dimensional Gaussian-type kernels were considered. This choice was shown to be good enough in these works to reproduce a large number of contrast- and orientation-dependent Poggendorff-like illusions via the LHE model in \eqref{eq:LHERef}, but not by the WC one \eqref{eq:WCRef}.

Here, motivated by he discussion in Section~\ref{sec:connectivity}, we study the effect of a more natural choice for the interaction kernel $\omega_\xi$, which we set as $\omega_\xi(\eta) = k_\tau(\xi,\eta)$, where $k_\tau:\mathcal M\times\mathcal M\to \mathbb R$ is the sub-Riemannian heat kernel evaluated at time $\tau>0$. Indeed, 3-dimensional isotropic Gaussian kernels are obtained via the Euclidean heat equation are not coherent with the intrinsically anisotropic neuronal connectivity structure of V1. Recalling \eqref{eq:evolution}, this choice of $\omega_\xi$ allows to rewrite the WC equations \eqref{eq:WCRef} as
\begin{equation}\tag{sR-WC}\label{eq:sr-wc}
    \partial_t a(\xi,t) = -(1+\lambda) a(\xi,t)+ \frac{1}{2M} e^{\tau\mathcal L}\left[\sigma\left( a(\cdot,t) \right)\right](\xi) + \lambda a_0(\xi)+\mu(\xi).
\end{equation}
Using this formulation, the evaluation of the interaction term at point $(\xi,t)\in\mathcal{M}\times (0,T]$ can be done by solving the sub-Riemannian heat equation and let it evolve for a  certain inner-time $\tau>0$. This avoids to deal directly with the explicit expression of $k_\tau$ whose numerical implementation is very delicate, as explained, e.g., in \cite{zhangNumerical2016}.

A similar simplification is not readily available for the LHE equation \eqref{eq:LHERef}, due to the dependence on $\xi$ of the integrand function. In this setting, we follow the discussion in \cite{bertalmio2007perceptual} and replace the non-linearity $\hat{\sigma}$ by a polynomial approximation of sufficiently large order $n$. Namely, we look for a polynomial approximation of $\hat\sigma$ of the form $\hat\sigma(r) = c_0 + \ldots + c_n r^n$, which allows us to write
\begin{equation}\label{eq:RTerm}
\begin{split}
\hat{\sigma}  \Big ( a(\xi,t)- a(\eta,t)  \Big ) & \approx   \sum_{i=0}^n\underbrace{\Big [ \sum_{j=0}^i (-1)^{j-i+1}c_j\begin{pmatrix} j \\ i   \end{pmatrix}a^{j-i}(\xi,t)\Big ]}_{ C_i(\xi, t):=} a^i(\eta,t) \\
 & = \sum_{i=0}^n C_i(\xi, t)a^i(\eta,t).
\end{split}
\end{equation}
This allows to approximate the interaction term in~\eqref{eq:LHERef} as
\begin{equation}\label{eq:approxFilteringLHE}
    \begin{split}
        \int_{Q\times [0,\pi)}k_\tau(\xi,\eta)\hat{\sigma}\Big( a(\xi,t) - a(\eta, t) \Big )d\eta 
        &\approx \sum_{i=0}^nC_i(\xi,t)\int_{Q\times [0,\pi)}k_\tau(\xi,\eta)a^i(\eta, t)\,d\eta\\ 
        &= \sum_{i=0}^nC_i(\xi,t) \: e^{\tau \mathcal{L}}\left[a^i(\cdot, t)\right](\xi).
    \end{split}
\end{equation}
Finally, the resulting (approximated) sub-Riemannian LHE equation reads:
\begin{equation}  \tag{sR-LHE}\label{eq:sr-lhe}
    \partial_t a(\xi,t) = -(1+\lambda) a(\xi,t)+ \frac{1}{2M} \sum_{i=0}^nC_i(\xi,t) \: e^{\tau\mathcal{L}}\left[a^i(\cdot, t)\right](\xi) + \lambda a_0(\xi)+\mu(\xi).
\end{equation}

\section{Discrete modelling and numerical realisation}\label{sec:Algorithms}

In this Section, we report a detailed description of how models \eqref{eq:sr-wc} and \eqref{eq:sr-lhe} can be formulated in a complete discrete setting, providing, in particular, some insights on how the sub-Riemannian evolution can be realised. We further add a self-contained section regarding the gradient-descent algorithm used to perform the numerical experiments reported in Section \ref{sec:experiments}, for more details see \cite{SSVMproceeding2019,bertalmio2020cortical}.

\subsection{Discrete modelling and lifting procedure via cake wavelets}
\label{sec:Discrete_Gabor_coeff}

The sub-Riemannian diffusion  $\operatorname{e}^{\tau \mathcal{L}}$ is discretised by a final time  $\tau = m\Delta\tau$ where $m$ and $\Delta\tau$ denote the number of iterations and the time-step, respectively. For $N\in\mathbb{N}^+$ and $\Delta x,\Delta y\in \R^+$ denoting the spatial sampling size, we then discretise the given gray-scale image function $f_0$ associated to the retinal stimulus on a uniform square spatial grid $Q:=\{ (x_i,y_j) = (i\Delta x, j\Delta y): i,j = 1,2,\dots,N\}\subset\mathbb{R}^2$ and denote, for each $i,j= 1,2,\dots,N$,  the brightness value at point $\zeta_{i,j}:=(x_i,y_j)\in Q$ by
\begin{equation}  \label{eq:discrete_f0}
F_0[i,j]=f_0(x_i, y_j)=f_0(\zeta_{i,j}).
\end{equation}
As far as the orientation sampling is concerned, we use a uniform orientation grid with points $\Theta:=\{\theta_k := k\Delta\theta, k=1,\dots, K\},~ K\in\mathbb{N}^+$ and $\Delta\theta=\pi/K$ .  We can then define the discrete version of the simple cell response $a_0(x_i, y_j, \theta_k)$ to the visual stimulus located at $\zeta_{i,j}\in Q$ with local orientation $\theta_k\in\Theta$ at time $t=0$ of the evolution as
\begin{equation}\label{eq:discrete-lift}
A_0[i,j,k]=a(x_i,y_j,\theta_k, 0) = a(\zeta_{i,j},\theta_k, 0) = (L f_0)_{i,j,k},
\end{equation}
where $L:Q\to Q\times \Theta$ is the lifting operation to be defined.

To do so, we consider in the following the image lifting procedure based on cake kernels introduced in \cite{duits2007invertible} and used, e.g., in \cite{bekkers2014multi,SSVMproceeding2019,bertalmio2020cortical}. We write the cake kernel centered at $\zeta_{i,j}$ and rotated by $\theta_k$ as
\begin{equation}
\Psi_{[i,j,k]}[\ell,m]=\psi_{(\zeta_{i,j},\theta_k)}(x_\ell, x_m) ,
\end{equation}
where $\ell,m\in\{1,2,\dots,N\}$. We can then write the lifting operation applied to the initial image $f_0$ for all $\zeta_{i,j}\in Q$ and $\theta_k\in\Theta$ as:
\begin{equation}  \label{eq:lifting_op}
(Lf_0)_{i,j,k}=A_0[i,j, k]=\sum_{l,m}\Psi_{[i,j,k]}[\ell,m]\,f_0[\ell,m].
\end{equation}

Finally, for $P\in\mathbb{N}^+$ we consider a time-discretisation of the interval $(0,T]$ at time nodes $\mathcal{T}:=\{t_p:=p\Delta t, p=1,\ldots P \}, P\in\mathbb{N}^+$ with $\Delta t:=T/P$. 

The resulting fully-discretised neuronal activation at $\zeta_{i,j}=(x_i,y_j)\in Q$, $\theta_k\in\Theta$ and $t_p \in \mathcal{T}$ will be thus denoted by:
\begin{equation}
A_p[i,j,k]=a(\zeta_{i,j},\theta_k, t_p).
\end{equation}

\subsection{Sub-Riemannian heat diffusion}\label{sec:sr-heat-discrete}

Let $g:\mathcal M\to\R$ be a given cortical stimulus, and denote  and set $G[i,j,k]=g(\xi_{i,j}, \theta_k)$. In this section we describe how to compute
\begin{equation}   \label{eq:sR_diff}
    \exp_\tau G[i,j,k] \approx e^{\tau\mathcal L}g(\zeta_{i,j},\theta_k),
\end{equation}
The main difficulty here is due the degeneracy arising from the anisotropy of the sub-Laplacian. Indeed, developing the computations in
\eqref{eq:heat-sr}, we have
\begin{equation}
    \mathcal L = \mathcal D^T \ell \mathcal D, \qquad 
    \mathcal D = 
    \left(\begin{array}{c}\partial_x\\ \partial_y \\ \partial_\theta\end{array}\right),
    \qquad
    \ell = \left(
    \begin{array}{ccc}
        \cos^2\theta & \cos\theta\sin\theta & 0 \\
        \cos\theta\sin\theta & \sin^2\theta & 0 \\
        0 & 0 & \beta^2
    \end{array}
    \right).
\end{equation}
In particular, it is straightforward to deduce that the eigenvalues of $\ell$ are $(0,\beta^2, 1)$.

The discretisation of such anisotropic operators can be done in several ways, see for example \cite{duits2010left1, duits2010left2, mirebeau2014anisotropic, baspinar2020sub}. In our implementation, we follow the method presented in \cite{boscainHypoelliptic2014} which is tailored around the  group structure of $\se$, the universal cover of $\mathcal M$, and based on the non-commutative Fourier transform, see also \cite{BCGPR18}.

It is convenient to assume for the following discussion $\Delta x = \Delta y = \sqrt{N}$ and $\Delta\theta = \pi/K$. 
The ``semi-discretised'' sub-Laplacian $\mathcal L_K$ can be defined by
\begin{equation}  \label{eq:semi_discrete_sR}
    \mathcal L g \approx \mathcal L_K G := D^2 G +\Lambda_K G,
\end{equation}
where by $\Lambda_K$ we denote the central difference operator discretising the derivatives along the $\theta$ direction, i.e. the operator
\begin{equation}
    \partial_\theta^2 G[i,j,k] \approx \Lambda_K G[i,j,k] = \frac{g(\xi_{i,j}, \theta_{k-1}) -2g(\xi_{ij}, \theta_{k}) +g(\xi_{i,j}, \theta_{k+1}) }{2}.
\end{equation}
The operator $D$ is the diagonal operator defined by
\begin{equation}
    D G[i,j,k] = \left(\cos (k\Delta\theta) \partial_x  + \sin(k\Delta\theta) \partial_y\right) g(\xi_{i,j}, \theta_k).
\end{equation}
The full discretisation is then achieved by discretising the spatial derivatives as
\begin{gather}
        \partial_x G[i,j,k] \approx \frac{\sqrt{N}}{2} \left(g(\xi_{i+1,j}, \theta_{k})-g(\xi_{i-1,j}, \theta_{k-1})\right),\\
    \partial_y G[i,j,k] \approx \frac{\sqrt{N}}{2} \left(g(\xi_{i,j+1}, \theta_{k})-g(\xi_{i,j-1}, \theta_{k-1})\right).
\end{gather}
 
Under the discretisation $\mathcal{L}_K$ of  $\mathcal{L}$ defined in \eqref{eq:semi_discrete_sR}, we now resort to Fourier methods to compute efficiently the solution of the sub-Riemannian heat equation
\begin{equation}  \label{eq:sR_heat_semidisc}
    \partial_t \psi = \mathcal L_g \psi,\qquad  \psi|_{t=0}=g.
\end{equation}
In particular, 
let $\hat G[r,s,k]$ be the discrete Fourier transform (DFT) of G w.r.t. the variables $i,j$, i.e. 
\begin{equation}
    \hat G[r,s,k] = \frac1N \sum_{r,s=1}^{N} G[i,j,k] e^{\frac{\iota 2\pi }N \left( (r-1)(i-1) + (s-1)(j-1) \right)}.
\end{equation}
A straightforward computation shows that 
\begin{equation}
\begin{split}
    \widehat{DG}[r,s,k] = & \iota \sqrt{N} d[r,s,k]\hat G[r,s,k],\\
    d[r,s,k] := & \cos (k\Delta\theta) \sin\left(\frac{2\pi r}N\right)  + \sin(k\Delta\theta)\sin\left(\frac{2\pi s}N\right).
    \end{split}
\end{equation}
Hence, \eqref{eq:sR_heat_semidisc} is mapped by the DFT to the following completely decoupled system of $N^2$ ordinary linear differential equations on $\mathbb C^K$:
\begin{equation}\label{eq:discr-sys}
    \begin{cases}
     \frac{d}{dt}  \Psi_t[r,s,\cdot] = \left(\Lambda_N - \frac{N}{2} \operatorname{diag}_k d[r,s,k]^2 \right)  \Psi_t[r,s,\cdot],\\
     \Psi_0[r,s,k] = \hat G[r,s,k]
    \end{cases}
     \qquad  r,s\in \{1,\ldots, N\},
\end{equation}
which can be solved efficiently through a variety of standard numerical schemes. We chose the semi-implicit Crank-Nicolson method \cite{crank1947practical} for its good stability properties. Let us remark that the operator at the r.h.s.\ of the above equations are periodic tridiagonal matrices, i.e., tridiagonal matrices with additional non-zero values at positions $(1,K)$ and $(K,1)$. Thus, the linear system appearing at each step of the Crank-Nicolson method can be solved in linear time w.r.t.\ $K$ via a variation of the Thomas algorithm.

The desired solution $\exp_\tau G$ can be then simply recovered by applying the inverse DFT to the solution of \eqref{eq:discr-sys} at time $\tau$.

\subsection{Discretisation via gradient descent}

We follow \cite{bertalmio2007perceptual, bertalmio2020cortical, bertalmio2020visual} and discretise both models \eqref{eq:sr-wc} and \eqref{eq:sr-lhe} via a simple explicit gradient descent scheme. Denoting the discretised version of the local mean average $\mu(\xi)$ appearing in the models by $U[i, j,k]= \mu(i\Delta x, y\Delta j, k\Delta\theta)$, we have that the the time stepping reads for all $p\geq 1$
\begin{equation}  \label{eq:grad_desc_discr}
A_{p}[i,j,k] =A_{p-1}[i,j,k]+ \Delta t \Big(-(1+\lambda)A_{p-1}[i,j,k] + A_0[i,j,k] + \lambda U[i,j,k]+ SA_{p-1}[i,j,k]\Big ),
\end{equation}
where $SA_{p-1}$ is defined depending on the model by:
\begin{equation}\label{eq:SA}
    SA_{p-1}[i,j,k] = \exp_\tau \sigma(A_{p-1})[i,j,k]
    \qquad\text{or}\qquad
    SA_{p-1}[i,j,k] = \sum_{\ell=0}^n C_{\ell,p-1}[i,j,k] \exp_\tau A_{p-1}[i,j,k],
\end{equation}
with $C_{\ell,p-1}$ being the discretised version of the coefficient $C_\ell$ in \eqref{eq:RTerm} at time $t_{p-1}$.

A sufficient condition on the time-step $\Delta t$ guaranteeing the convergence of the numerical scheme \eqref{eq:grad_desc_discr} is $\Delta t\leq 1/(1+\lambda)$ (see \cite{bertalmio2007perceptual}).

\subsection{Pseudocode}

Our algorithmic procedure consists of three main numerical sub-steps. The first one is the lifting of the two dimensional input image $f_0$ to the space $\mathcal{M}$ via \eqref{eq:lifting_op}. The second one is the Fourier-based procedure described in Section \ref{sec:sr-heat-discrete} to compute the sub-Riemannian diffusion \eqref{eq:sR_diff} which can be used as kernel to describe the neuronal interactions along the horizontal connection. This step is intrinsically linked to the last iterative procedure, based on computing the gradient descent update \eqref{eq:grad_desc_discr}-\eqref{eq:SA} describing the evolution of neuronal activity in the cortical framework both for the  \eqref{eq:sr-wc} and  \eqref{eq:sr-lhe}.

We report the simplified pseudo-code below. The detailed Julia package used to produce the following examples is freely available at the following webpage \url{https://github.com/dprn/srLHE}.

\begin{algorithm*}[H]
\SetAlgoLined
\KwData{Initial image $f_0[i,j]$ \\ \textbf{Parameters}: $\lambda, \alpha, \sigma_\mu, \alpha, K, \beta, \Delta t, T, M $, \texttt{tol}}
\KwResult{Processed image $F[i,j]$}
 Compute lift $A_0[i,j,k] \leftarrow Lf_0[i,j,k]$ via \eqref{eq:discrete-lift}\;
 Initialize iteration index $p\leftarrow 0$\;
 \Repeat{${\|A_p - A_{p-1}\|}/{\|A_p\|} < \texttt{tol}$}{
  $p\leftarrow p+1$\;
  Compute 
  interaction term $SA_{p-1}$ via \eqref{eq:SA}\;
  Compute $A_{p}$ via GD update \eqref{eq:grad_desc_discr}\;
 }
 Projection on retinal plane $F[i,j] \leftarrow \sum_{k=1}^K A_p[i,j,k]$\;
 \caption{sR-WC and sr-LHE pseudocode}
 \label{algo:pseudocode}
\end{algorithm*}

\section{Numerical Experiments}  \label{sec:experiments}

\begin{figure}[th!]
\begin{subfigure}[b]{0.45\textwidth}
\centering
 \includegraphics[height=5cm]{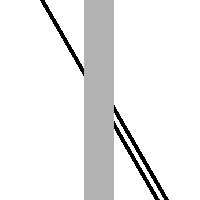}
\caption{Poggendorf illusion.}
\label{fig:pogg_test}
\end{subfigure}
\begin{subfigure}[b]{0.45\textwidth}
\centering
\includegraphics[height=5cm]{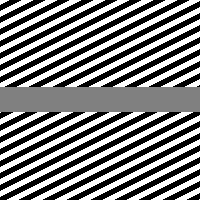}
\caption{Poggendorff gratings.}
\label{fig:pogg2}
\end{subfigure}
\caption{Gray-scale Poggendorff-type illusions. Figure \eqref{fig:pogg_test} is the standard $200\times 200$ Poggendorff illusion with a 30 pixel-wide central and an incidence angle of $\pi/3$ drawn by the black lines with the central bar .
Figure \eqref{fig:pogg2} is a variation of the classical Poggendorff illusion where a further background grating is present.}
\label{fig:Poggendorff_tests}
\end{figure}

In this section we present the results obtained by applying models \eqref{eq:sr-wc},  \eqref{eq:sr-lhe} via Algorithm \ref{algo:pseudocode} to two Poggendorf-type illusions reported in Figure~\ref{fig:Poggendorff_tests}. Our results are compared to the  ones obtained by applying the corresponding  WC and LHE 3-dimensional models with a 3D-Gaussian kernel as described in \cite{bertalmio2020cortical,bertalmio2020visual}.  The objective of the following experiments is to understand whether the output produced by applying \eqref{eq:sr-wc} and \eqref{eq:sr-lhe} to the images in Figure~\ref{fig:Poggendorff_tests} agrees with the illusory effects perceived. Since the quantitative assessment of the strength of these effects is a challenging problem, the outputs of Algorithm \ref{algo:pseudocode} have too be evaluated by visual inspection. Namely, for each output, we consider whether the continuation of a fixed black stripe on one side of a central bar connects with a segment on the other side. Differently from inpainting-type problems, we stress that for these problems the objective is to replicate the perceived wrong alignments due to contrast and orientation effects rather than its collinear prosecution and/or to investigate when both type of completions can be reproduced.

\textbf{Testing data: Poggendorff-type illusions.} We test the \eqref{eq:sr-wc} and  \eqref{eq:sr-lhe} models on a grayscale version of the Poggendorff illusion in Figure~\ref{fig:Poggendorff} and on its modification reported in Figure~\ref{fig:pogg2} where the background is constituted by a grating pattern: in this case the perceived bias depends also on the contrast between the central surface and the background lines.

\textbf{Parameters.} Images in Figure \ref{fig:Poggendorff_tests} have size $N \times N$ pixels, with $N=200$. The lifting procedure to the space of positions and orientations is obtained by discretising $[0,\pi)$ into  $K=16$ orientations (this is in agreement with the standard range of 12-18 orientations typically considered to be relevant in literature \cite{Chariker2016,Pattadkal2018}). The relevant cake wavelets are then computed following \cite{bekkers2014multi}, setting the frequency band $\texttt{bw}=5$ for all experiments. The scaling parameter $\beta$ appearing in \eqref{eq:heat-sr} is set\footnote{Such parameter adjusts the different spatial and orientation sampling. A single spatial unit is equal to $\sqrt{2}$ pixel edge whereas a single orientation unit is $1$ pixel edge.} to 
$\beta = {K}/(N^2\sqrt{2})$, and the parameter $M$ appearing in  \eqref{eq:sr-wc}, \eqref{eq:sr-lhe} is set to $M=1$. 

Parameters varying from test to test are: the slope $\alpha>0$ of the sigmoid functions $\sigma$  in \eqref{eq:sigma} and $\hat{\sigma}$, the fidelity weight $\lambda>0$, the variance of the 2D Gaussian filtering $\sigma_\mu$ use to compute the local mean average $\mu$  in \eqref{eq:sr-wc} and \eqref{eq:sr-lhe}, the gradient descent time-step $\Delta t$, the time step $\Delta \tau$ and the final time $\tau$ 
used to compute the sub-Riemannian heat diffusion $\operatorname{e}^{\tau \mathcal{L}}$.

\subsection{Poggendorff gratings}   

\begin{figure}[th]
\centering
\begin{subfigure}[b]{.45\textwidth}
\centering
    \includegraphics[height=5cm]{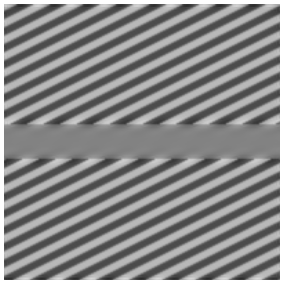}
    \caption{(WC)}
    \label{fig:gratings-WCold}
\end{subfigure}
\begin{subfigure}[b]{.45\textwidth}
\centering
    \includegraphics[height=5cm]{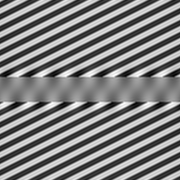}
    \caption{(sR-WC)}
    \label{fig:gratings-srWC}
\end{subfigure}
\caption{Model output for Poggendorff gratings in Figure \ref{fig:pogg2} via WC models. Figure~\ref{fig:gratings-WCold}: result of the WC model proposed in \cite{bertalmio2020cortical,bertalmio2020visual}. Figure~\ref{fig:gratings-srWC}: result of \eqref{eq:sr-wc} with parameters $\lambda=0.01$, $\alpha = 20$, $\sigmamu = 6.5$, $\Delta t=0.1$, $\Delta\tau=0.01$, $\tau = 5$.}
\label{fig:WC-gratings}
\end{figure}

In Figure \ref{fig:WC-gratings} we report the results obtained by applying \eqref{eq:sr-wc} to the Poggendorff grating image in Figure~\ref{fig:pogg2}. We compare them with the ones obtained by the cortical-inspired WC model considered \cite{bertalmio2020cortical,bertalmio2020visual} where the sR heat-kernel is an isotropic 3D Gaussian which are reported in Figure \ref{fig:gratings-WCold}.
In~Figure~\ref{fig:gratings-srWC}, we observe that the sR diffusion encoded in \eqref{eq:sr-wc} favours the propagation of the grating throughout the central gray bar so that the resultant image agrees with our perception of misalignment. We stress that such illusion could not be reproduced via the cortical-inspired isotropic WC model proposed in \cite{bertalmio2020cortical,bertalmio2020visual}. The use of the appropriate sub-Laplacian diffusion is thus crucial in this example to replicate the illusion. 

\begin{figure}[t]
\centering
\begin{subfigure}[b]{.45\textwidth}
\centering
    \includegraphics[height=5cm]{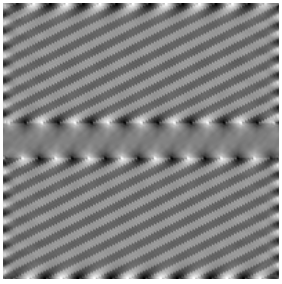}
    \caption{(LHE)}
    \label{fig:gratings-LHEold}
\end{subfigure}
\begin{subfigure}[b]{.45\textwidth}
\centering
    \includegraphics[height=5cm]{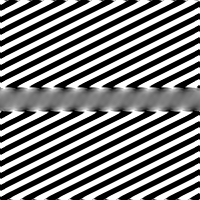}
    \caption{(sR-LHE)}
    \label{fig:gratings-srLHE}
\end{subfigure}
\caption{Model output for Poggendorff gratings in Figure \ref{fig:pogg2} via LHE models. Figure~\ref{fig:gratings-LHEold}: result of the LHE model proposed in \cite{bertalmio2020cortical,bertalmio2020visual}. Figure~\ref{fig:gratings-srLHE}: result of \eqref{eq:sr-lhe} with parameters $\alpha=8$, $\tau=5$, $\lambda=2$, $\sigmamu = 1$, $\Delta t=0.15$, $\Delta\tau=0.01$.
}
\label{fig:LHE-gratings}
\end{figure}

We further report in Figure \ref{fig:LHE-gratings} the result obtained by applying \eqref{eq:sr-lhe} on the same image.
We observe that in this case both \eqref{eq:sr-lhe} model and the LHE cortical model introduced in \cite{bertalmio2020cortical,bertalmio2020visual} reproduce the illusion. 

Note that both \eqref{eq:sr-wc} and \eqref{eq:sr-lhe} further preserve fidelity w.r.t.~the given image outside the target region, which is not the case in the LHE cortical model presented in \cite{bertalmio2020cortical,bertalmio2020visual}.

\subsection{Dependence on parameters: inpainting vs. perceptual completion}

The capability of the 
\eqref{eq:sr-lhe} model to reproduce visual misperceptions depends on the chosen parameters. 
This fact was already observed in \cite{bertalmio2020visual} for the cortical-inspired LHE model proposed therein endowed by a standard Gaussian filtering. There, LHE was shown to reproduce illusory phenomena only in the case where the chosen standard deviation of the Gaussian filter was set to be large enough (w.r.t.\ the overall size of the image). On the contrary, the LHE model was shown to perform geometrical completion (inpainting) for small values of the standard deviation. Roughly speaking, this corresponds to the fact that perceptual phenomena -- such as geometrical optical illusions -- can be modelled only when the interaction kernel is wide enough for the information to cross the central gray line. This is in agreement with psycho-physical experiences in \cite{Weintraub1971} where the width of the central missing part of the Poggendorff illusion is shown to be directly correlated with the intensity of the illusion.

\begin{figure}[htb!]
\centering
\begin{subfigure}[b]{.3\textwidth}
\centering
    \includegraphics[width=.9\textwidth]{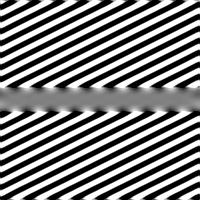}
\caption{($\tau=0.1$)}
\label{fig:inp-srlhe-1}
\end{subfigure}
\begin{subfigure}[b]{.3\textwidth}
\centering
    \includegraphics[width=.9\textwidth]{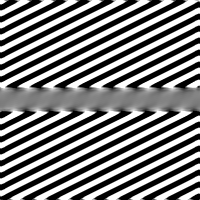}
\caption{($\tau=0.5$)}
\label{fig:inp-srlhe-2}
\end{subfigure}
\begin{subfigure}[b]{.3\textwidth}
\centering
    \includegraphics[width=.9\textwidth]{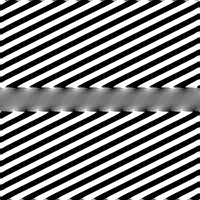}
\caption{($\tau=2.5$)}
\label{fig:inp-srlhe-3}
\end{subfigure}
\caption{Sensitivity to the parameter $\tau$ for \eqref{eq:sr-lhe} model for the visual perception of Figure \ref{fig:pogg2}. The completion inside the central gray bar changes from geometrical (inpainting type) to illusory (perception type). Parameters: $\tau$ varies from $0.1$ to $5$, $\alpha=6$, $\lambda=2$, $\sigmamu = 1$, $\Delta t=0.15$, $\Delta\tau=0.01$.
}
\label{fig:LHE_inpainting}
\end{figure}

In the case under consideration here, the parameter encoding the width of the interaction kernel is the final time $\tau$ of the sub-Riemannian diffusion used to model the activity propagation along neural connections.
To support this observation, in Figure~\ref{fig:LHE_inpainting} we show that the completion obtained via \eqref{eq:sr-lhe} shifts from a geometrical one (inpainting), where $\tau$ is small, to a perceptual one, for $\tau$ sufficiently big.

As far as \eqref{eq:sr-wc} model is concerned, we observed that despite the improved capability of replicating the Poggendorf gratings, the transition from perceptual completion to inpainting could not be reproduced. In agreement with the efficient representation principle, this supports the idea that visual perceptual phenomena are better encoded by variational models as \eqref{eq:sr-lhe} than by non-variational ones as \eqref{eq:sr-wc}.

\subsection{Poggendorff illusion}

\begin{figure}[ht]
\centering
\begin{subfigure}[b]{.45\textwidth}
\centering
    \includegraphics[height=5cm]{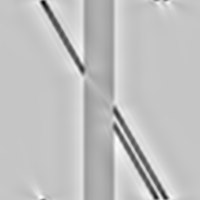}
    \caption{(LHE)}
    \label{fig:orig-LHEold}
\end{subfigure}
\begin{subfigure}[b]{.45\textwidth}
\centering
    \includegraphics[height=5cm]{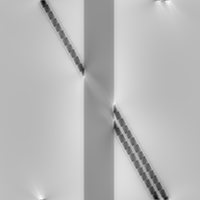}
    \caption{(sR-LHE)}
    \label{fig:orig-srLHE}
\end{subfigure}

\medskip
\begin{subfigure}[b]{.45\textwidth}
\centering
    \includegraphics[height=3cm]{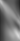}
    \caption{(LHE), zoomed.}
    \label{fig:orig-LHEold-z}
\end{subfigure}
\begin{subfigure}[b]{.45\textwidth}
\centering
    \includegraphics[height=3cm]{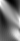}
    \caption{(sR-LHE), zoomed}
    \label{fig:orig-srLHE-z}
\end{subfigure}
\caption{Model output for Poggendorff illusion in Figure \ref{fig:pogg_test} via LHE models. Figure~\ref{fig:orig-LHEold}: result of the LHE model proposed in \cite{bertalmio2020cortical,bertalmio2020visual} (with parameters $\sigma_\mu=2$, $\sigma_\omega=12$, $\lambda=0.7$, $\alpha= 5$). Figure~\ref{fig:orig-srLHE}: result of \eqref{eq:sr-lhe} with parameters $\alpha=8$, $\tau=2.5$, $\lambda=0.5$, $\sigmamu = 2.5$, $\Delta t=0.15$, $\Delta\tau=0.1$. 
Figure~\ref{fig:orig-srLHE-z} (resp.\ Figure~\ref{fig:orig-LHEold-z}):  zoom and renormalization on $[0,1]$ of the central region of the result in Figure~\ref{fig:orig-srLHE} (resp.~Figure~\ref{fig:orig-LHEold}). 
}
\label{fig:pogg_orig_LHE}
\end{figure}
In Figure \ref{fig:pogg_orig_LHE} we report the results obtained by applying LHE methods to the standard Poggendorff illusion in Figure Figure~\ref{fig:pogg_test}. In particular, in Figure~\ref{fig:orig-LHEold} we show the result obtained via the LHE method of \cite{bertalmio2020cortical,bertalmio2020visual}, while in Figure~\ref{fig:orig-srLHE} we show the result obtained via \eqref{eq:sr-lhe}, with two close-ups in Figures \ref{fig:orig-LHEold-z} and \ref{fig:orig-srLHE-z} showing a normalized detail of the central region onto the set of values $[0,1]$.
As shown by these preliminary examples, the prosecutions computed by both (LHE) models agree with our perception as the reconstructed connection in the target region links the two misaligned segments, while somehow 'stopping' the connection of the collinear one. 

This phenomenon as well as a more detailed study on how the choice of the parameters used to generate Figure \ref{fig:pogg_test} (such as the incidence angle, the width of the central gray bar, the distance between lines), in a similar spirit to \cite{Retsa2020} where psysho-physicis experiments were performed on analogous images, is an interesting topic for future research.

\FloatBarrier

\section{Conclusion}

In this work we presented a sub-Rimannian version of the Local Histogram Equalization mean-field model previously studied in \cite{bertalmio2020cortical, bertalmio2020visual} and here denoted by \eqref{eq:sr-lhe}.  The model considered is a natural extension of existing ones where 
 the  kernel used to model neural interactions was simply chosen to be 3D Gaussian kernel, while in \eqref{eq:sr-lhe} this is chosen as the sub-Riemannian kernel of the heat equation formulated in the space of positions and orientations given by the primary visual cortex (V1). A numerical procedure based on Fourier expansions is described to compute such evolution efficiently and in a stable way and a gradient-descent algorithm is used for the numerical discretisation of the model

We tested the \eqref{eq:sr-lhe} model on orientation-dependent Poggendorff-type illusions and showed that (i) in presence of a sufficiently wide interaction kernel, model \eqref{eq:sr-lhe} is capable to reproduce the perceptual misalignments perceived, in agreement with previous work (see Figures~\ref{fig:LHE-gratings} and \ref{fig:pogg_orig_LHE}), (ii) when the interaction kernel is too narrow, (sr-LHE)  favors a geometric-type completion (inpainting) of the illusion (see Figure~\ref{fig:LHE_inpainting}) due to the limited amount of diffusion considered.

We also considered \eqref{eq:sr-wc}, a similar model obtained by using the sub-Riemannian interaction kernel in the standard Wilson-Cowan equations. We showed that the introduction of such cortical-based kernel improves the capability of WC-type models of reproducing Poggendorff-type illusions, in comparison to the analogous results reported \cite{bertalmio2020cortical,bertalmio2020visual} where the cortical version of WC with a standard 3D Gaussian kernel was shown to fail to replicate the illusion.

Finally, we stress that, in agreement with the standard range of 12-18 orientations typically considered to be relevant in literature \cite{Chariker2016,Pattadkal2018}, all the aforementioned results have been obtained by considering $K=16$ orientations. The LHE and WC models previously proposed were unable to obtain meaningful results with less that $K=30$ orientations. 

\section*{Acknowledgments}
LC, VF and DP acknowledge the support of a public grant overseen by the French National Research Agency (ANR) as part of the \textit{Investissement d'avenir program}, through the iCODE project funded by the IDEX Paris-Saclay, ANR-11-IDEX-0003-02. VF acknowledges the support received from the European Union's Horizon 2020 research and innovation programme under the Marie Sklodowska-Curie grant agreement No 794592.

\bibliographystyle{ieeetr}
\bibliography{meetingNotes_Bib}

\end{document}